\documentclass[letterpaper, 10 pt, conference]{styles/ieeeconf}
\IEEEoverridecommandlockouts                     

\usepackage{url}
\usepackage{color}
\usepackage[dvipsnames]{xcolor}
\usepackage{wrapfig}
\usepackage{booktabs}
\usepackage[font={footnotesize}]{subcaption}
\usepackage[font={footnotesize}]{caption}

\usepackage{amsmath}
\usepackage{float}
\usepackage{optidef}
\usepackage{setspace}
\usepackage{graphicx}
\usepackage{colortbl}
\usepackage{mathrsfs}
\usepackage{amssymb}
\usepackage{accents}

\usepackage{nicefrac}
\usepackage{algorithm}
\usepackage{dirtree}
\usepackage{cases}
\usepackage[noend]{algpseudocode}

\usepackage{flushend}
\usepackage{multicol}

\usepackage{pifont}
\newcommand{\cmark}{\textcolor{ForestGreen}{\ding{51}}}%
\newcommand{\xmark}{\textcolor{red}{\ding{55}}}%

\makeatletter
\newcommand\fs@spaceruled{\def\@fs@cfont{\bfseries}\let\@fs@capt\floatc@ruled
  \def\@fs@pre{\vspace{0.4\baselineskip}\hrule height.8pt depth0pt \kern2pt}%
  \def\@fs@post{\vspace{-0.4\baselineskip}\kern2pt\hrule\relax\vspace{-12pt}}%
  \def\@fs@mid{\kern2pt\hrule\kern2pt}%
  \let\@fs@iftopcapt\iftrue}
\makeatother

\usepackage{lipsum}
\usepackage{pgfplots}
\pgfplotsset{tick label style={font=\bfseries\color{white!15!black}},}
\pgfplotsset{compat=newest}
\newlength\figH
\newlength\figW

\DeclareMathOperator*{\argmin}{arg\,min}
\DeclareMathOperator{\proj}{proj}

\usepackage{cite}
\definecolor{darkgreen}{rgb}{0.0,0.45,0.0}

\usepackage{hyperref}
\hypersetup{
  colorlinks = true,
  urlcolor   = blue,
  linkcolor  = blue,
  citecolor  = darkgreen
} 

\definecolor{melon}{rgb}{0.99, 0.74, 0.71}
\definecolor{my_gray}{HTML}{616B85}

\usepackage{listings}

\definecolor{dkgreen}{rgb}{0,0.6,0}
\definecolor{gray}{rgb}{0.5,0.5,0.5}
\definecolor{mauve}{rgb}{0.58,0,0.82}

\usetikzlibrary{plotmarks}

\lstdefinelanguage{Julia}%
  {morekeywords={abstract,break,case,catch,const,continue,do,else,elseif,%
      end,export,false,for,function,immutable,import,importall,if,in,%
      macro,module,otherwise,quote,return,switch,true,try,type,typealias,%
      using,while},%
   sensitive=true,%
   alsoother={\$},%
   morecomment=[l]\#,%
   morecomment=[n]{\#=}{=\#},%
   morestring=[s]{"}{"},%
   morestring=[m]{'}{'},%
}[keywords,comments,strings]%

\newcommand{\algName}{Conic-TinyMPC}

\newcommand{\algNameCodeDir}{tinympc}

\title{\LARGE \bf Code Generation and Conic Constraints for Model-Predictive\\Control on Microcontrollers with \algName}

\author{Ishaan Mahajan$^{1*}$, Khai Nguyen$^{2,3*}$, Sam Schoedel$^{2*}$, 
        Elakhya Nedumaran$^{2}$, Moises Mata$^{1}$, \\
        Brian Plancher$^{4,5}$, and Zachary Manchester$^{2,3}$
\thanks{This material is based upon work supported by the National Science Foundation (under Award 2411369). Any opinions, findings, conclusions, or recommendations expressed in this material are those of the authors and do not necessarily reflect those of the funding organizations.}%
\thanks{$^{1}$ School of Engineering and Applied Science, Columbia University}%
\thanks{$^{2}$ Carnegie Mellon University}%
\thanks{$^{3}$ Massachusetts Institute of Technology}%
\thanks{$^{4,5}$ Barnard College, Columbia University and Dartmouth College}%
\thanks{$*$Equal Contribution. Correspondence to: {\tt\scriptsize plancher@dartmouth.edu}}%
}

\begin{document}
\maketitle
\thispagestyle{empty}
\pagestyle{empty}


\begin{abstract}
    Model-predictive control (MPC) is a state-of-the-art control method for constrained robotic systems, yet deployment on resource-limited hardware remains difficult. This challenge is magnified by expressive conic constraints, which offer greater modeling power but require significantly more computation than linear alternatives.
    To address this challenge, we extend recent work developing fast, structure-exploiting, cached solvers for embedded applications based on the Alternating Direction Method of Multipliers (ADMM) to provide support for second-order cones, as well as \texttt{C++} code generation from \texttt{Python}, \texttt{MATLAB}, and \texttt{Julia}. Microcontroller benchmarks show that our solver provides up to a two-order-of-magnitude speedup, ranging from 10.6x to 142.7x, over state-of-the-art embedded solvers on QP and SOCP problems, and enables us to fit order-of-magnitude larger problems in memory. We validate our solver's deployed performance through simulation and hardware experiments, including trajectory tracking with conic constraints on a 27g Crazyflie quadrotor. Our open-source code is available at \url{https://tinympc.org}.
\end{abstract}

\section{Introduction} \label{sec:intro}
Model Predictive Control (MPC) is an algorithmic approach that enables highly dynamic online control for robots subject to actuator and state constraints~\cite{wensing2022optimization,aydinoglu2023consensus,le2024fast, gurumurthy2025deq}. However, while MPC has been deployed quite successfully in both academia and industry, its application is often hindered by computational limitations. This challenge is amplified when dealing with tiny, low-cost, low-power robots, as their onboard microcontroller units (MCUs) feature orders-of-magnitude less RAM, flash memory, and processor speed compared to the CPUs and GPUs available on larger robots~\cite{tinympc,neuman2022tiny,zhang2017visual}. Consequently, many examples of intelligent behaviors executed on these tiny platforms rely on off-board compute~\cite{lambert2019low,luis2020online,torrente2021data,xi2021gto,AMSwarm}. For deployment on such limited computational platforms, which also often lack full hardware support for floating-point arithmetic, an ideal MPC solver should be division-free, use only static memory allocation, and support warm starting to take advantage of computation at previous time steps~\cite{marcucci2020warm, adabag2024mpcgpu, altroc, reluqp}. Compiled code should also have a low memory footprint and be easily verifiable through an interface to a high-level language (e.g., \texttt{Python}). 

Moreover, while many embedded solvers today focus solely on quadratic programming (QP), second-order cone programs (SOCPs) represent a significantly richer class of tractable convex optimization problems, strictly generalizing linear and quadratic ones while remaining efficiently solvable due to their symmetric, self-dual structure~\cite{nesterov1994self,Boyd04}. Many problems that appear nonconvex admit exact or lifted SOCP reformulations~\cite{alizadeh2003second,ben2009robust}. Such problems also arise naturally in robotic and aerospace control problems involving friction, attitude, and thrust limits~\cite{lobo1998applications,liu2016entry,klein1990optimal}. As such, an ideal embedded MPC solver should natively support SOCPs while efficiently addressing their computational demands.

\begin{figure}[t!]
    \centering
    \includegraphics[width=0.8\columnwidth,trim={0cm 0.3cm 0 0.3cm},clip]{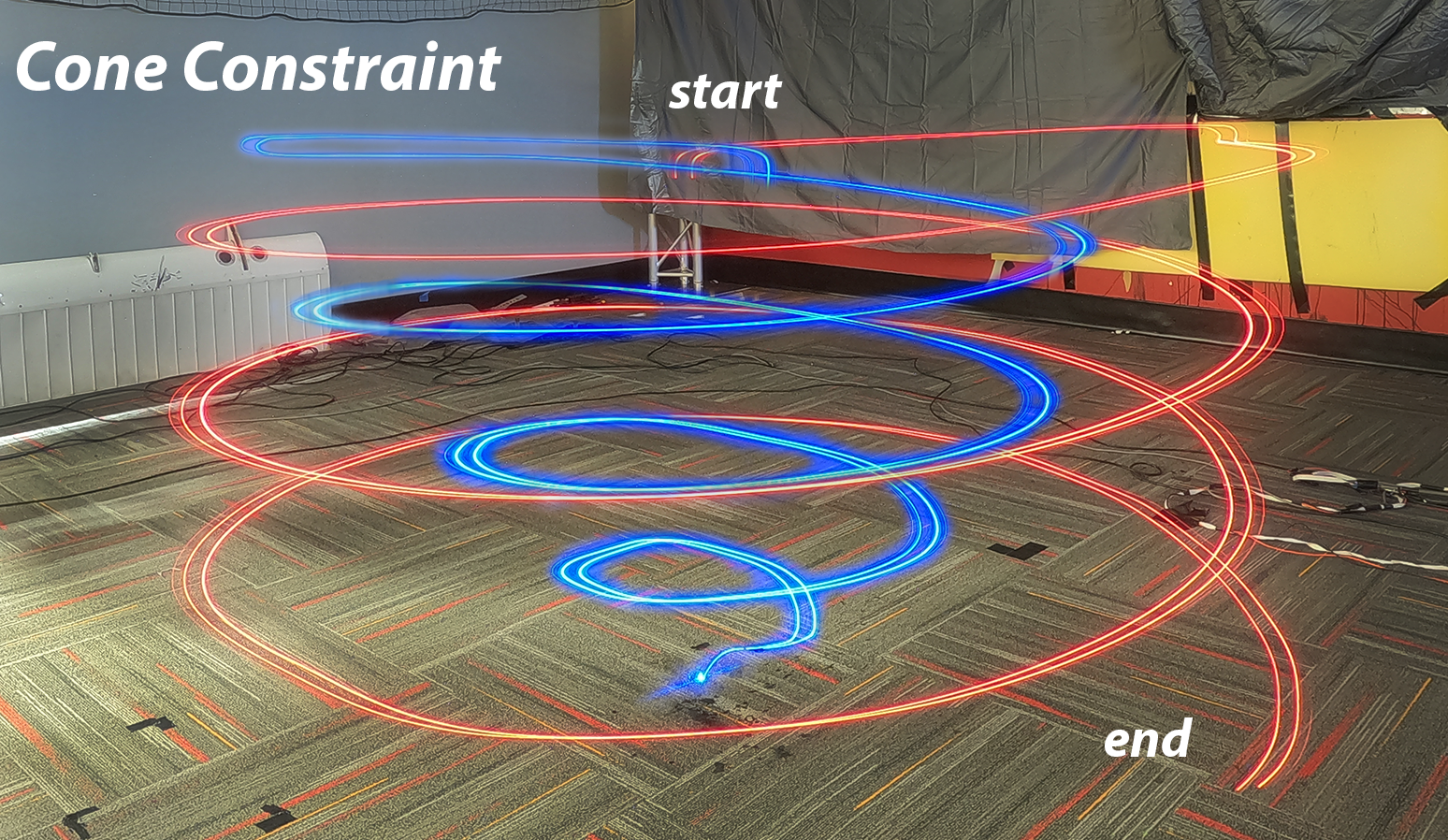}\\
    \vspace{1pt}
    \includegraphics[width=0.8\columnwidth]{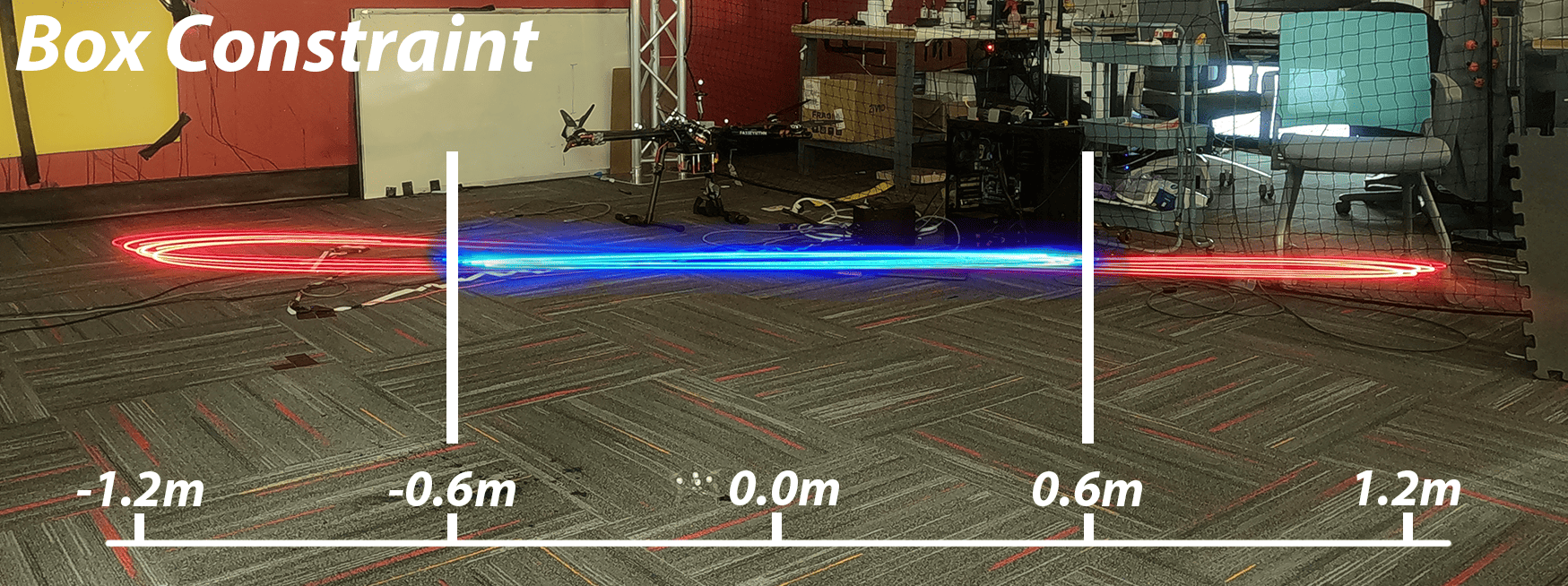}
    \caption{
    We demonstrate our solver using a 27 gram nano quadrotor, the Crazyflie.
    \textbf{Top:} we track a descending helical reference (\textcolor{red}{red}) with its position subject to a 45$^{\circ}$ second-order cone glideslope. This requires the aircraft to perform a spiral landing maneuver (\textcolor{blue}{blue}). \textbf{Bottom:} we design a predictive safety filter to guarantee safe maneuvers within a box-shaped space (\textcolor{blue}{blue}) regardless of the nominal controller behavior (\textcolor{red}{red}).}
    \label{fig:hw_landing_safe}
    \vspace{-20pt}
\end{figure}
\begin{table*}[t!]
    \centering
    \vspace{5pt}
    \caption{Comparison of general-purpose and model predictive control solvers. 
    \vspace{-4pt}
    \label{tab:solver_spec}}
    \small
    \setlength\extrarowheight{-0.5pt}
    \begin{tabular}{ccccccc}
        \toprule
        \textbf{Solver} & \textbf{SOC} & \textbf{Warm Starting} & \textbf{Embedded} & \textbf{Open Source} & \textbf{Quad. Obj.} & \textbf{MPC Tailored}\\
        \midrule
        Clarabel \cite{clarabel}        & \cmark & \xmark & \xmark & \cmark & \cmark  & \xmark \\
        COSMO \cite{cosmo}              & \cmark & \cmark & \xmark & \cmark & \cmark  & \xmark \\
        ECOS \cite{ecos}                & \cmark & \xmark & \cmark & \cmark & \xmark  & \xmark \\
        MOSEK \cite{mosek}              & \cmark & \xmark & \xmark & \xmark & \cmark  & \xmark \\
        OSQP \cite{stellato2020osqp}    & \xmark & \cmark & \cmark & \cmark & \cmark  & \xmark \\
        SCS \cite{scs}                  & \cmark & \cmark & \cmark & \cmark & \cmark  & \xmark \\
        \midrule
        FORCESPRO \cite{forces}            & \xmark & \cmark & \cmark & \xmark & \cmark  & \cmark \\
        ALTRO-C \cite{altroc}           & \cmark & \cmark & \xmark & \cmark & \cmark  & \cmark \\
        acados \cite{Verschueren2021}   & \xmark & \cmark & \xmark & \cmark & \cmark  & \cmark \\
        HPIPM \cite{frison2020hpipm}   & \xmark & \cmark & \xmark & \cmark & \cmark  & \cmark \\
        TinyMPC \cite{tinympc}   & \xmark & \cmark & \cmark & \cmark & \cmark  & \cmark \\
\rowcolor[gray]{.95} \algName (ours)     & \cmark & \cmark & \cmark & \cmark & \cmark  & \cmark \\
        \bottomrule
    \end{tabular}
    \vspace{-10pt}
\end{table*}

Table~\ref{tab:solver_spec} compares commonly used SOCP and QP solvers to illustrate how well they align with these design criteria. Several efficient optimization solvers and techniques suitable for embedded MPC have emerged in recent years~\cite{megahertz2014, Donoghue2013}, with notable software packages including OSQP~\cite{stellato2020osqp}, CVXGEN~\cite{Mattingley12}, ECOS~\cite{ecos}, and SCS~\cite{scs}. However, because many of these are not purpose-built for MPC, they either do not easily support warm starting, don't take advantage of problem or sparsity structure, are not designed to easily enable embedded deployment, or some combination of these issues. In contrast, while TinyMPC~\cite{tinympc} is the first MPC solver tailored for dynamic tiny robot control on MCUs, it (as well as OSQP and CVXGEN) only supports QPs. TinyMPC also lacked a convenient high-level programming interface.

To address these shortcomings, in this work, we develop \algName. Our contributions include: 1) \textit{support for conic constraints}, focusing on SOCPs (Section~\ref{sec:algorithm}), a critical need for many real-world robotics applications; and 2) an \textit{open-source code generation software package} with \texttt{Python}, \texttt{MATLAB}, and \texttt{Julia} interfaces to both ease the deployment of embedded MPC, as well as provide code generation and solution verification examples (Section~\ref{sec:codegen}). 

We present microcontroller benchmarks (Section~\ref{ss:benchmark}) demonstrating up to a two-order-of-magnitude speedup and one-order of-magnitude reduction in memory usage over state-of-the-art embedded QP and SOCP solvers. We also validate our solver’s deployed performance through hardware experiments on a 27g Crazyflie quadrotor (Section~\ref{ss:hardwareExp}), including trajectory tracking with conic constraints.
Our open-source code is available at \url{https://tinympc.org}.

\section{Background} \label{sec:background}
\subsection{The Linear-Quadratic Regulator}

The linear-quadratic regulator (LQR) problem~\cite{lewis12optimal} is an optimal control problem in which a quadratic cost function is minimized subject to linear (or affine) dynamics constraints:
\begin{equation} \label{eq:lqr}
\begin{split}
    \min_{\substack{x_{1:N}, u_{1:N-1}}} 
        & J = \tfrac{1}{2}x_N^{\intercal} Q_N x_N + q_N^{\intercal} x_N + \\ 
        \sum_{k=1}^{N-1} &\tfrac{1}{2}x_k^{\intercal} Q_k x_k + q_k^{\intercal} x_k + \tfrac{1}{2}u_k^{\intercal} R_k u_k + r_k^{\intercal} u_k\\
        \textrm{subject to} \quad & x_{k+1} = A_k x_{k} + B_k u_{k} + c_k, \; \forall k \in [1,N),\\
\end{split}
\end{equation}
\noindent where $x_k \in \mathbb{R}^n$, $u_k \in \mathbb{R}^m$ are the state and control at time step $k$, $N$ is the number of time steps, 
$A_k \in \mathbb{R}^{n \times n}$, $B_k \in \mathbb{R}^{n \times m}$, and $c_k \in \mathbb{R}^n$ define the system dynamics, $Q_k \succeq 0$, $R_k \succ 0$, and $Q_N \succeq 0$ are symmetric cost-weighting matrices and $q_k$ and $r_k$ are linear cost vectors.
Equation~\eqref{eq:lqr} is a classical problem in the field of optimal control whose solution is an affine feedback controller \cite{lewis12optimal}:
\begin{equation}\label{eq:lqrSolution} 
    u_k^* = -K_k x_k - d_k.
\end{equation}
Feedback and feedback terms ($K_k$, $d_k$) are found by solving the discrete-time Riccati equation backward in time, starting with $P_N = Q_N$ and $p_N = q_N$, where $P_k$ and $p_k$ are the quadratic and linear terms of the cost-to-go function \cite{lewis12optimal}:\\
\begin{equation}\label{eq:riccati}
\scalebox{0.9}{$
\begin{split}
K_k &= (R_k+B_k^\intercal P_{k+1}B_k)^{-1} (B_k^\intercal P_{k+1}A_k)\\
d_k &= (R_k+B_k^\intercal P_{k+1}B_k)^{-1} (B_k^\intercal p_{k+1} + r_k + B_k^\intercal P_{k+1}c_k)\\
P_k &= Q_k + K_k^\intercal R_k K_k + (A_k-B_kK_k)^\intercal P_{k+1} (A_k-B_kK_k)\\
p_k &= q_k + (A_k-B_kK_k)^\intercal (p_{k+1}-P_{k+1}B_kd_k + P_{k+1}c_k) \\ &\qquad\; + K_k^\intercal(R_kd_k - r_k).    
\end{split}
$}
\end{equation}


\subsection{Convex Model-Predictive Control}

Convex MPC extends this to admit additional convex constraints on the states and controls (as shown in \textcolor{blue}{blue}):
\begin{equation}\label{eq:convexMpc}
\begin{aligned}
    \min_{x_{1:N},u_{1:N-1}} \quad & J(x_{1:N},u_{1:N-1})\\
\textrm{subject to} \quad & x_{k+1} = A_k x_{k} + B_k u_{k} + c_k, \; \forall k \in [1,N) \\
& \textcolor{blue}{x_k \in \mathcal{X}, u_k \in \mathcal{U}, \quad \forall k \in [1,N)},\\
\end{aligned}
\end{equation}
where $\mathcal{X}$ and $\mathcal{U}$ are convex sets. The convexity of this problem means that it can be solved efficiently and reliably, enabling real-time deployment in a variety of control applications, including autonomous rocket landings~\cite{13rocketlanding}, legged locomotion~\cite{18legged}, and autonomous driving~\cite{18driving}.

When $\mathcal{X}$ and $\mathcal{U}$ can be expressed as linear constraints, \eqref{eq:convexMpc} is a QP. When $\mathcal{X}$ and $\mathcal{U}$ can be expressed as both linear and second-order cone constraints, \eqref{eq:convexMpc} is an SOCP, and can be put into the standard form (where $\mathcal{K}$ is a cone):
\begin{equation}\label{eq:socp}
\begin{aligned}
    \min_{x \in \mathbb{R}^n} \quad &  \tfrac{1}{2}x^{\intercal}Px + q^{\intercal}x \\
\textrm{subject to} \quad & Gx \leq h, \; \textcolor{blue}{x \in \mathcal{K}}.
\end{aligned}
\end{equation}
The addition of the final constraints in \textcolor{blue}{blue} separate the SOCP from the QP. Further analysis, including feasibility and stability guarantees can be found in~\cite{Boyd04,scs}.


\subsection{Alternating Direction Method of Multipliers (ADMM)}

We provide a very brief summary of ADMM here and refer readers to~\cite{boyd2011distributed} for more~details. Given a generic optimization problem (with $f$ and $\mathcal{C}$ convex):
\begin{equation}\label{eq:simpleOpt}
\begin{aligned}
\min_{x} \quad & f(x)\\
\textrm{subject to} \quad & x \in \mathcal{C},
\end{aligned}
\end{equation}
we can form the equivalent problem, introducing slack $z$, and indicator function $I_{\mathcal{C}}$:
\begin{equation}\label{eq:simpleOptTransformed}
\begin{aligned}
\min_{x, z} \quad &f(x) + I_{\mathcal{C}}(z) \\
\textrm{subject to} \quad &x = z
\end{aligned}
\quad
I_{\mathcal{C}}(z) = \begin{cases}
       0 & z \in \mathcal{C}  \\
       \infty & \text{otherwise}.
    \end{cases} 
\end{equation}
The augmented Lagrangian of the transformed problem \eqref{eq:simpleOptTransformed} is (with Lagrange multiplier $\lambda$ and scalar penalty weight~$\rho$):
\begin{equation}\label{eq:simpleOptTransformedAugL}
    \mathcal{L}_A(x,z,\lambda) = f(x) + I_{\mathcal{C}}(z) + \lambda^{\intercal}(x-z) + \tfrac{\rho}{2}||x-z||_2^2.
\end{equation}

If we perform alternating minimization of~\eqref{eq:simpleOptTransformedAugL} with respect to $x$ and $z$, we arrive at the three-step ADMM iteration,\vspace{-2pt}
{%
\begin{align}
    \text{primal update}: x^{+} &= \argmin_x \mathcal{L}_A(x,z,\lambda) ,\label{eq:primal_update}\\ \vspace{-8pt}
    \text{slack update}: z^{+} &= \argmin_z \mathcal{L}_A(x^{+},z,\lambda),\label{eq:slack_update}\\ \vspace{-8pt}
    \text{dual update}: \lambda^{+} &= \lambda + \rho (x^+ - z^+) \label{eq:dual_update},
\end{align}%
}\vspace{-2pt}%
where the last step is a gradient-ascent update on the Lagrange multiplier~\cite{boyd2011distributed}. These steps can be iterated until a desired convergence tolerance is achieved. 

In the special cases of QPs and SOCPs, each step of the ADMM algorithm becomes very simple to compute: the primal update is the solution to a linear system, the slack update is a linear or conic projection, and the dual update is simply scaled vector addition. As such, the computational complexity of the three steps for QPs and SOCPs is:
\begin{itemize}
    \item $\mathcal{O}(n^3)$ for the primal update~\eqref{eq:primal_update},
    \item $\mathcal{O}(n^2)$ for the slack update~\eqref{eq:slack_update},
    \item and $\mathcal{O}(n)$ for the dual update~\eqref{eq:dual_update}.
\end{itemize}
Due to this simplicity, ADMM-based QP and SOCP solvers have demonstrated state-of-the-art results~\cite{stellato2020osqp,scs}.


\subsection{TinyMPC}
TinyMPC~\cite{tinympc}, exploits properties of the MPC problem through pre-computation and caching with an ADMM framework to efficiently solve this problem via three assumptions:
\begin{enumerate}
    \item The dynamical system can be modeled as linear time invariant, with fixed $A,B,c\;\forall k \in [0,N)$;
    \item The quadratic cost can be modeled with fixed hessians, $Q,R \;\forall k \in [0,N)$, $Q_N$; and
    \item The finite horizon LQR feedback gain and cost-to-go Hessian, $K_k, P_k$, can be effectively approximated by the solution to the infinite-horizon LQR solution, $K_{\text{inf}}, P_{\text{inf}} \;\forall k \in [0,N]$.
\end{enumerate}

We provide a brief summary of the approach and refer to~\cite{tinympc} for more details.
TinyMPC~\cite{tinympc} splits the standard LQR problem from all additional state and input constraints via ADMM. The primal update, \eqref{eq:primal_update}, becomes: 
\begin{equation}\label{eq:lqr_primal}
\begin{aligned}
    \min_{x_{1:N},u_{1:N-1}} \quad & \frac{1}{2}x_N^{\intercal} \tilde{Q}_N x_N + \tilde{q}_N^{\intercal}x_N + \\
    \sum_{k = 1}^{N-1} & \frac{1}{2}x_k^{\intercal}\tilde{Q}x_k + \tilde{q}_k^{\intercal}x_k
    + \frac{1}{2}u_k^{\intercal}\tilde{R}u_k + \tilde{r}^{\intercal}u_k\\
\textrm{subject to} \quad & x_{k+1} = A x_{k} + B u_{k} + c,\\
\end{aligned}
\end{equation}
where we use scaled dual variables $y$ and $g$ and for convenience~\cite{boyd2011distributed} and the following are defined:
\begin{equation}\label{eq:lin_quad_updated_cost}
    \begin{aligned}
        &\tilde{Q}_N = Q_N + \rho I, \quad 
        \tilde{Q} = Q + \rho I, \quad
        \tilde{R} = R + \rho I, \\
        &\tilde{q}_N = q_N + \rho(\lambda_N / \rho -  z_N) = q_N + \rho(y_N -  z_N),\\
        &\tilde{q}_k = q_k + \rho(\lambda_k / \rho - z_k) = q_k + \rho(y_k - z_k),\\
        &\tilde{r}_k = r_k + \rho(\mu_k / \rho -  w_k) = r_k + \rho(g_k -  w_k).\\
    \end{aligned}
\end{equation}
This enables a slight simplification to \eqref{eq:dual_update}, where we substitute the dual variables with their scaled forms and eliminate $\rho$. As a result, the scaled dual variables $y_k$ and $g_k$ are always equal to the difference between the primal and slack variables, which is a convergence criteria that now does not need to be recalculated during convergence checks.

As \eqref{eq:lqr_primal} has the same form as \eqref{eq:lqr}, it can be solved efficiently through a backwards Riccati recursion, \eqref{eq:riccati}, followed by an affine dynamics roll-out of the resulting policy, \eqref{eq:lqrSolution}.
The slack update remains a projection onto the feasible set:
\begin{equation}\label{eq:mpc_slack}
    \begin{aligned}
    z^{+}_k = \proj_\mathcal{X} (x^+_k+y_k), \\
    w^{+}_k = \proj_\mathcal{U} (u^+_k+g_k),
    \end{aligned}
\end{equation}
where the superscript denotes the variable at the subsequent ADMM iteration, and the dual update becomes:
\begin{equation}
    \begin{aligned}\label{eq:mpc_dual}
    y^{+}_k = y_k + x^+_k -  z^+_k, \\
    g^{+}_k = g_k + u^+_k -  w^+_k .
    \end{aligned}
\end{equation}

Given a long enough horizon, the Riccati recursion \eqref{eq:riccati} converges to the solution of the infinite-horizon LQR problem \cite{lewis12optimal}. \cite{tinympc} exploits this property and assumes that the single infinite horizon gain, $K_{\text{inf}}$, and cost-to-go Hessian, $P_{\text{inf}}$, sufficiently approximate the time-varying values, $K_k, P_k$. Combining this approximation with our assumption of fixed $A,B,c,Q,Q_N,R$ matrices enables us to drastically simplify the Riccati recursion, \eqref{eq:riccati}, not only easing its computational complexity, but also greatly reducing its memory footprint as we only need to cache $A,B,c,Q,Q_N,R,K_{\text{inf}},P_{\text{inf}}$, along with a handful of other precomputed and cached constants:
\begin{equation}\label{eq:cache}
    \begin{aligned}
        C_1 &= (R + B^\intercal P_\text{inf}B)^{-1}, \\
        C_2 &= (A-BK_\text{inf})^\intercal, \\
        C_3 &= C_1B^\intercal P_{\text{inf}}c, \\
        C_4 &= C_2 P_{\text{inf}}c. \\
    \end{aligned}
\end{equation}
Using these terms, the LQR backward pass simplifies to:
\begin{equation}\label{eq:fast_riccati}
    \begin{aligned}
        d_k &= C_1(B^\intercal p_{k+1} + r_k) + C_3, \\
        p_k &= q_k + C_2p_{k+1} - K_\text{inf}^\intercal r_k + C_4,\\
    \end{aligned}
\end{equation}
which only requires matrix-vector products to compute, reducing computational complexity of the primal update from $\mathcal{O}(n^3)$ to $\mathcal{O}(n^2)$, drastically reducing online computation time, and avoiding online division entirely.
We note that $C_3$ and $C_4$ are derived in addition to $C_1$ and $C_2$ from~\cite{tinympc} to support dynamics with the additional constant term $c$.

Finally, we note that ADMM solvers like OSQP~\cite{stellato2020osqp} adaptively scale the penalty term $\rho$ in \eqref{eq:simpleOptTransformedAugL} for performance. However, this requires performing additional matrix factorizations. To avoid this,~\cite{tinympc,reluqp} pre-compute and cache sets of matrices corresponding to several values of $\rho$, which we refer to as the set $[\varrho]$. Online, the solver switches between these values of $\rho$, and their respective cached matrices, based on the values of the primal and dual residuals using heuristics adapted from OSQP~\cite{stellato2020osqp}.\footnote{We also note that recent work~\cite{Mahajan2025adaptiveTinyMPC} proposes additional schemes to adapt $\rho$ with finer-grained updates. We will integrate this advance into our open-source conic framework in future work.}

\section{The \algName~Solver} \label{sec:algorithm}
As noted in~\cite{tinympc}, the slack update in \eqref{eq:slack_update} can be expressed as the operator $\Pi$, which projects the slack variable onto its feasible set. More generally, this projection step can be defined for any convex set. Because the ADMM algorithm naturally isolates the projection subproblem, any convex set with a computationally efficient projection operator can be seamlessly incorporated into our framework. Conveniently, many standard convex cones admit simple closed-form projection operators~\cite{boyd2011distributed}. We demonstrate this by example in the remainder of this section. 

We first note that projection onto a linear inequality constraint, or equivalently, projection of a point $z$ to a hyperplane $\mathcal{H} = \{x:\langle x,a \rangle=b\}$, can be written as follows: 
\begin{equation}   \label{eq:proj_linear2}
   \begin{aligned}
       \Pi_{\mathcal{H}}(z) = \proj_{\mathcal{H}} (z) = z - \frac{\langle z,a \rangle-b}{||a||^2} a.
   \end{aligned}
\end{equation}
For constant bounds on variables, such as the case of position, velocity, or control limits, $(l,u)$, this can be reduced to a projection onto a set of upper and lower bounds:
\begin{equation}\label{eq:proj_linear}
    \begin{aligned}
        \Pi_{l,u}(z) &= \max(l, \min(u, z)).
    \end{aligned}
\end{equation}
 
This projection approach extends to conic problems in the same manner. We can, for example, define the second-order cone (``ice-cream cone'')~\cite{Boyd04} as follows:
\begin{equation}
    \mathcal{K} = \Bigl\{ z\in \mathbb{R}^n | z_n \geq \sqrt{z_1^2+z_2^2+\dots+z_{n-1}^2} \Bigr\},
\end{equation}
The second-order cone also admits a closed-form and compact projection operator:
\begin{equation}\label{eq:proj_cone}
    \begin{aligned}
        \Pi_{\mathcal{K}}(z) =
        \begin{cases}
        0,                & \|v\|_2 \leq -a, \\
        z,                & \|v\|_2 \leq a, \\
        \cfrac{1}{2}\Bigl(1+\cfrac{a}{\|v\|_2}\Bigr) \begin{bmatrix} v \\ \|v\|_2 \end{bmatrix},         & \|v\|_2 > |a|,
        \end{cases}
    \end{aligned}
\end{equation}
where $v = [z_1, \ldots, z_{n-1}]^\intercal$ and $a=z_n$. Here, $z_i, i=1,...,n$ is any vector subset of the state or control slack variables.
In principle, other cones can also be implemented, e.g. the cone of $n \times n$ positive semi-definite matrices (``semi-definite cone'')~\cite{Boyd04}.
Algorithm~\ref{alg:tinympc_alg_words} summarizes the overall algorithm.

\section{Code Generation} \label{sec:codegen}
To enable the community to more easily leverage \algName, we have developed a code-generation tool with \texttt{Python}, \texttt{MATLAB}, and \texttt{Julia} interfaces that produces dependency-free \texttt{C++} code for easy deployment. 
We hope that through such interfaces, and our additional examples, available alongside our open-source code, the community can quickly prototype and deploy our solver onto their tiny robot systems.
\begin{algorithm}[!t]
\caption{\algName}\label{alg:tinympc_alg_words}
\begin{algorithmic}
\Function{offline\underline{{ }{ }}precompute}{input}
    \State \textbf{for} $\rho \in [\varrho]$ \textbf{form} cache via \eqref{eq:lin_quad_updated_cost}, $K_{inf}, P_{inf}$, \eqref{eq:cache}
\EndFunction
\Function{online\underline{{ }{ }}solve}{input}
    \State \text{Select}~$\rho \in [\varrho]$ \text{and associated cached terms}
    \While {\text{not converged}}
    \State \texttt{//Primal Update}
    \State $ p_{1:N-1}, d_{1:N-1} \gets \text{Backward pass via \eqref{eq:fast_riccati}}$
    \State $x_{1:N}, u_{1:N-1} \gets \text{Forward pass via \eqref{eq:lqrSolution}}$
    
    \State \texttt{//Slack and Dual Updates}
    \State $z_{1:N}, w_{1:N-1} \gets \text{Projection via \eqref{eq:proj_linear2}, \eqref{eq:proj_linear}, or \eqref{eq:proj_cone}}$
    
    \State $y_{1:N}, g_{1:N-1}  \gets \text{Gradient ascent \eqref{eq:mpc_dual}}$
    \State $q_{1:N}, r_{1:N-1}, p_N \gets \text{Update linear cost terms}$
    \EndWhile
    
    \Return $x_{1:N}, u_{1:N-1}$
\EndFunction
\end{algorithmic}
\end{algorithm}
\begin{lstlisting}[float=!t, label={lst:gen_code}, language=Python, caption={A minimal Python script to generate MPC problem code.}]
import @\algNameCodePkg@
# Create the solver object
solver = @\algNameCodePkg@.@\algNameCodeObj@()
# Initialize the solver
solver.setup(N, A, B, c, Q, R, bnds, socs, options)
# Generate code 
solver.codegen(output_dir)
\end{lstlisting}
\begin{lstlisting}[float=!t, label={lst:verify_code}, language=Python, caption={An example Python script to run the generated code.}]
import numpy as np
import @\algNameCodePkgGen@
# Set initial state
@\algNameCodePkgGen@.set_x0(np.array([0.5, 0, 0, 0]))
# Solve the problem
solution = @\algNameCodePkgGen@.solve()
# Get the solution
controls = solution["controls"]
\end{lstlisting}
\begin{lstlisting}[float=!t, label={lst:gen_main}, language=C++, caption={A simple C++ program that loads the problem data from \texttt{tiny\_data\_workspace.hpp} and solves the problem.}]
#include "@\textcolor{mauve}{\algNameCodeDir}@.hpp"
#include "tiny_data_workspace.hpp"
int main(int argc, char **argv) {
    tiny_solve(&solver);  // Solve the problem
    return 0;
}
\end{lstlisting}
\begin{lstlisting}[label={lst:modify}, float=!t, language=C++, caption={Directly updating parameters of the MPC problem in C++.}, float=!h]
// Update initial/feedback state
tiny_set_x0(&solver, x0_new);
// Update trajectory reference
tiny_set_x_ref(&solver, xref_new);
// Update Bounds
tiny_set_bound_constraints(&solver, xmin_new, xmax_new, umin_new, umax_new);
\end{lstlisting}
\begin{figure}[!t]
    \footnotesize
    \vspace{-5pt}
    \DTsetlength{0.2em}{1em}{0.2em}{0.4pt}{1.6pt}
    \setlength{\DTbaselineskip}{8.5pt}
    \dirtree{%
    .1 <proj\underline{{ }}dir>.
    .2 include.
    .3 Eigen.
    .2 \algNameCodeDir.
    .3 [*.hpp].
    .3 [*.cpp].
    .2 src.
    .3 tiny\underline{{ }}data\underline{{ }}workspace.cpp.
    .3 tiny\underline{{ }}main.cpp.
    }
    \caption{The tree structure of the generated code. The main program is stored in \texttt{tiny\underline{{ }}main.cpp}.}
    \label{fig:dir_tree}
    \vspace{-20pt}
\end{figure}
In the remainder of this section, we describe our code-generation interfaces through examples and code listings using our \texttt{Python} interface and note that the process is nearly identical in \texttt{MATLAB} and \texttt{Julia}.

Listing~\ref{lst:gen_code} shows how to generate problem-specific code. The \texttt{setup} function initializes the problem with specific data, namely: time horizon ($N$), system model ($A$, $B$, and $c$), cost weights ($Q$ and $R$), linear and conic constraint parameters (\texttt{bnds} and \texttt{socs}), and solver \texttt{options}. For example, users may set primal and dual tolerances, or the maximum number of ADMM iterations. This kind of parameter tuning is often critical for returning a usable solution within real-time limits for particular systems of interest. The \texttt{codegen} function is then used to generate the custom-tailored code.

Users may choose to compile code for their host system either manually or through our interface for ease of testing when cumbersome to build in \texttt{C++}. Listing~\ref{lst:verify_code} shows an example script that loads the generated code library, solves the problem, then retrieves the solution. The reference trajectory and initial state may be set using \texttt{set\_x\_ref}, \texttt{set\_u\_ref}, and \texttt{set\_x0}, and may be done on the microcontroller using the \texttt{C++} equivalents. Additional wrapped functions exist for overwriting constraint parameters.

The directory structure of the resulting generated \texttt{C++} code is shown in Fig.~\ref{fig:dir_tree}. The solver's source code and associated headers are in the \texttt{\algNameCodeDir} subdirectory. The generated code is compact and does not rely on dynamic memory allocation, making it particularly suitable for embedded use cases. An example program is located in \texttt{tiny\_main.cpp}. This program imports workspace data from the \texttt{tiny\_data\_workspace.hpp} header and then solves the given problem (Listing~\ref{lst:gen_main}).

We also offer functions to update the initial state, reference trajectories, and constraints on the states and inputs using wrapper functions, which are essential in MPC settings (see Listing~\ref{lst:modify} for a number of examples).

We note that the \texttt{Python} interface not only allows users to generate \texttt{C++} code that may be run on a microcontroller, but it also allows the user to run TinyMPC functions directly in \texttt{Python}. This enables users to investigate the solver in a desktop environment before switching to a microcontroller. In future work, we also hope to build on these \texttt{Python} interfaces to enable us to build a complete \texttt{MicroPython} library, for even easier use on microcontrollers. Finally, we remind the reader that similar features, functions, and interfaces exist through our \texttt{MATLAB} and \texttt{Julia} interfaces.

\section{Experiments} \label{sec:results}
We benchmark the performance of the generated code from \algName~through a sets of microcontroller benchmarks and control tasks running onboard a 27 gram Crazyflie quadrotor~\cite{crazyflie2_1} to demonstrate TinyMPC's effectiveness in real-world deployed conditions. All experiments are available with our open-source code for reproducibility.

\subsection{Microcontroller Benchmarks}\label{ss:benchmark}

\subsubsection{Predictive Safety Filtering}
We first formulate a QP with box constraints on states and controls to act as a predictive safety filter for a nominal task policy as in~\cite{hsu2023safety, bejarano2023multi}. We compare the solution times and memory usage of \algName~against the state-of-the-art OSQP~\cite{stellato2020osqp} QP solver, utilizing both solvers' \texttt{Python} code generation interfaces, while varying the state and horizon dimensions. We benchmark on a STM32F405 Adafruit Feather board, which has an ARM Cortex-M4 operating at 168~MHz with 1~MB of flash memory and 128~kB of RAM, very similar to computational hardware on the Crazyflie 2.1 used for our later hardware experiments in Section~\ref{ss:hardwareExp}.

\begin{figure*}[!t]
\centering
\captionsetup[subfigure]{justification=centering,singlelinecheck=false}
\newlength{\subfigH}\setlength{\subfigH}{7.8cm} 
\begin{subfigure}[t]{0.495\textwidth}
    \centering
    \raisebox{0pt}[\subfigH][0pt]{%
        \includegraphics[width=\linewidth,height=\subfigH,keepaspectratio]{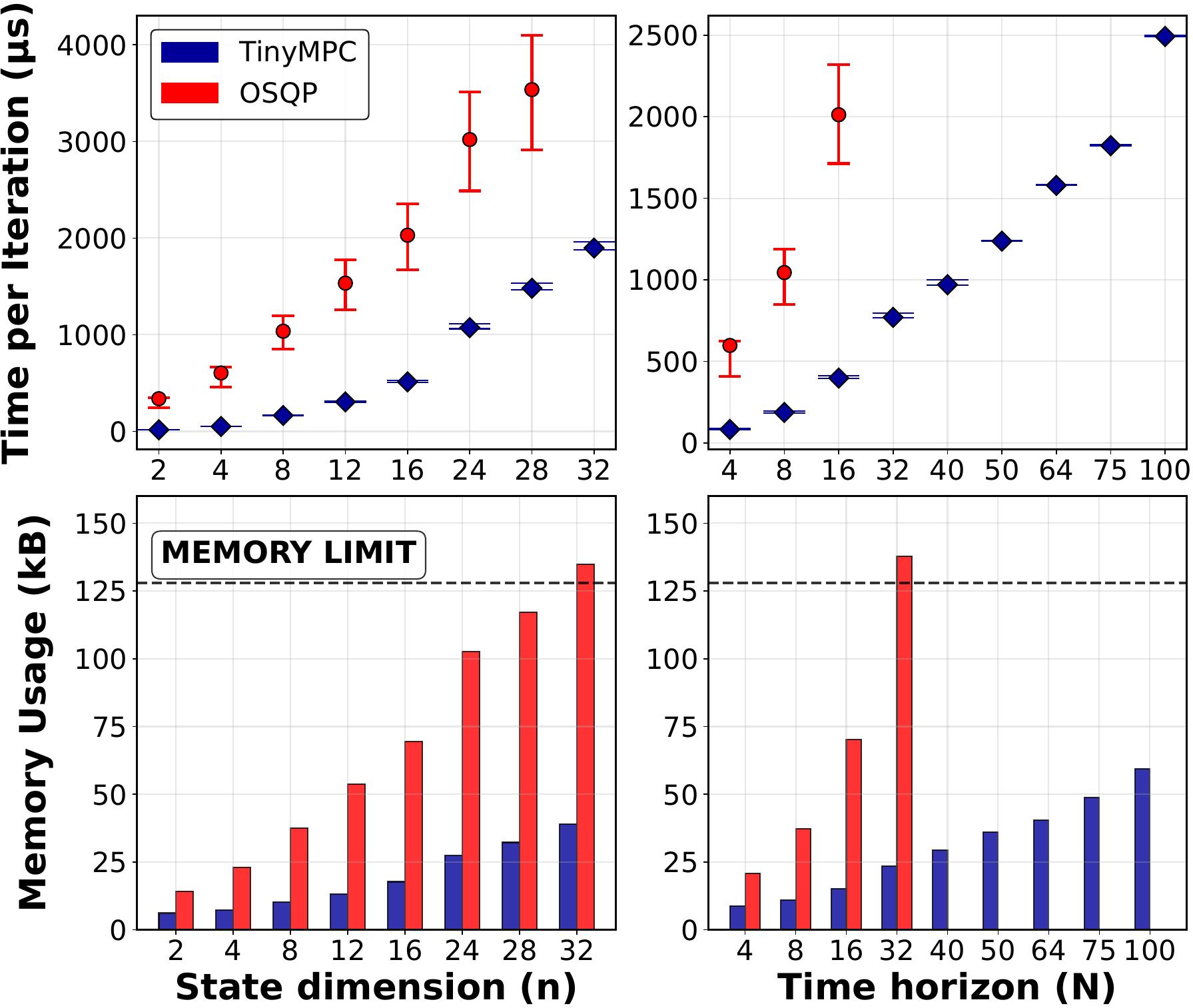}%
    }
    \caption{Predictive Safety Filtering}
    \label{fig:safety_filter}
\end{subfigure}\hfill
\begin{subfigure}[t]{0.495\textwidth}
    \centering
    \raisebox{0pt}[\subfigH][0pt]{%
        \includegraphics[width=\linewidth,height=\subfigH,keepaspectratio]{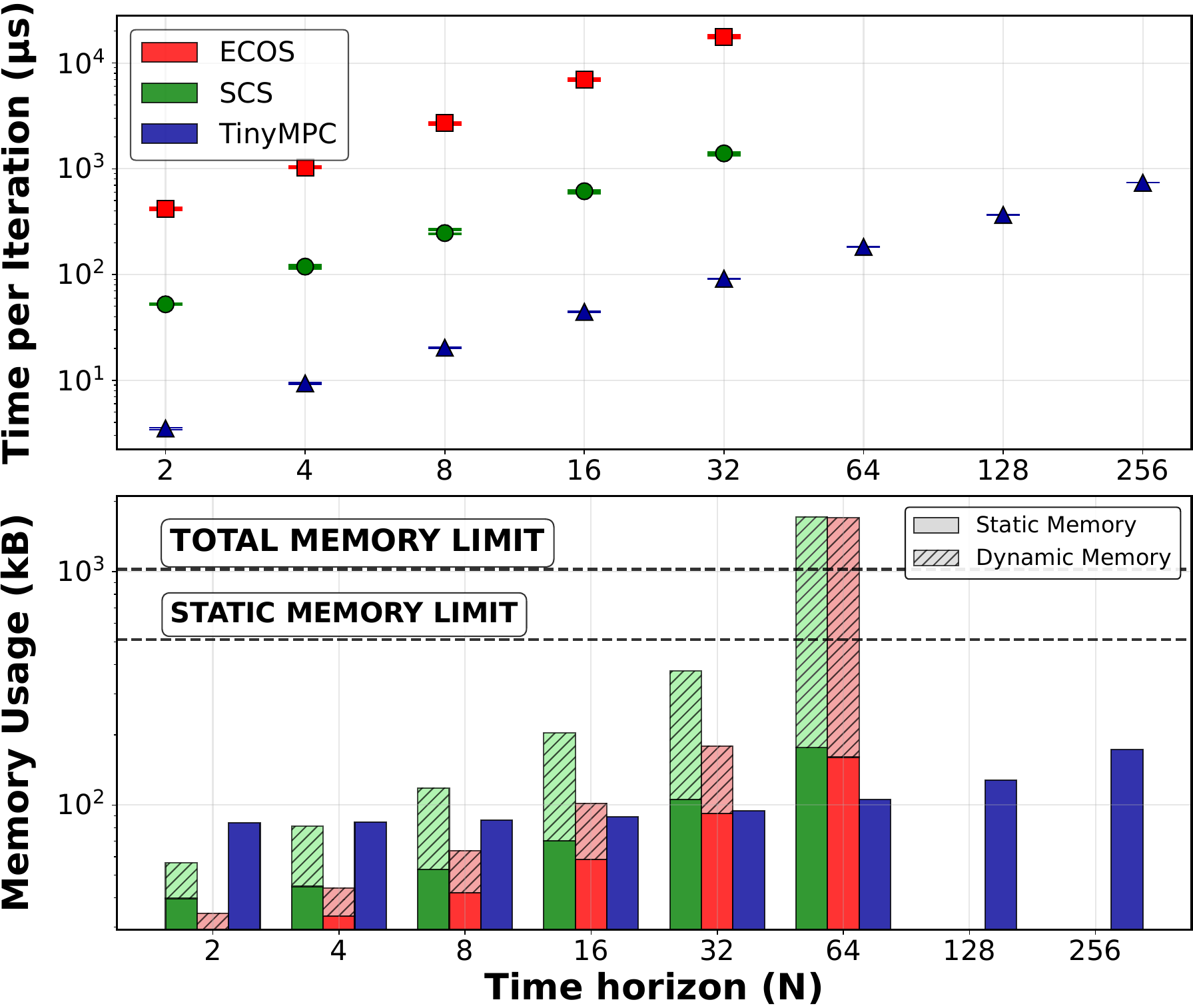}%
    }
    \caption{Rocket Soft Landing}
    \label{fig:rocket_landing}
\end{subfigure}
\caption{(a) Predictive safety filtering performance comparison between \algName~and OSQP on an STM32F405 Feather board. Top row shows average iteration times, bottom row shows memory usage. Left column: time horizon kept constant at $N=10$ while state dimension $n$ ranged from 2 to 32 and input dimension was set to half of the state dimension. Right column: state and control input held constant at $n=10$ and $m=5$ while $N$ ranged from 4 to 100. Error bars represent maximum and minimum time taken per iteration for all MPC steps. Black dotted lines denote memory thresholds. (b) Rocket soft-landing performance comparison between \algName, ECOS, and SCS using a Teensy 4.1 development board. Top plot shows memory usage, bottom plot shows average iteration times. In this SOCP-based experiment $n=6$ and $m=3$ while $N$ varied from 2 to 256. Error bars represent maximum and minimum time taken per iteration for all MPC steps performed. Black dotted lines denote memory thresholds.}
    \vspace{-15pt}
\end{figure*}

Fig.~\ref{fig:safety_filter} shows the total program size and the average execution times per iteration. \algName~uses drastically less memory and exhibits significant speed-ups over OSQP. For varying states, \algName~achieves up to 20.4× faster execution, while for varying time horizons, it achieves up to 7.2× faster execution.
Moreover, the reduction in memory usage allows \algName~to solve real-time optimal control of complex systems with long time horizons. In particular, \algName~was able to handle time horizons of up to 100 knot points, whereas OSQP surpassed the 128~kB memory capacity of the STM32 at a time horizon of only $N = 32$. Additionally, \algName~demonstrated scalability to larger state dimensions up to $n=32$, whereas OSQP encountered memory limitations beyond~$n = 28$.


\subsubsection{Rocket Soft-Landing} \label{ss:rocketSoftLanding}
The second benchmark is a rocket soft-landing problem which requires a rocket to land with small final velocity at a desired position, resulting in a conic glide-scope constraint. 
We benchmark the performance of \algName~again via it's \texttt{Python} code generation against ECOS~\cite{ecos} and SCS~\cite{scs}, state-of-the-art SOCP solvers, using CVXPYgen's~\cite{cvxpygen} code generation interface. All solver options were set to equivalent values wherever possible and all tolerances were set to $0.01$.

Here we benchmark on a Teensy 4.1~\cite{teensy41} development board, which has an ARM Cortex-M7 microcontroller operating at 600 MHz, with 7.75 MB of flash memory, 512 kB of tightly coupled static RAM, and an additional 512 kB of tightly coupled dynamic RAM. The increased compute and memory capacity of the Teensy was particularly important to enable us to benchmark against ECOS and SCS, and enabled us to collect more overall data as the largest SOCP problem involved 2301 decision variables as well as 1530 linear equality constraints, 1530 linear inequality constraints, and 255 second-order cone constraints. However, we note that \algName, even for this larger problem, could still fit on the more constrained Adafruit Feather used in the prior benchmark, as well as on the constrained MCU found on the Crazyflie 2.1, which we demonstrate via our hardware experiments in Section~\ref{ss:hardwareExp}.

Fig.~\ref{fig:rocket_landing} shows the amount of statically and dynamically allocated memory and the average execution times per iteration for varying time horizon. \algName~outperforms SCS and ECOS in execution time and memory, achieving an average speed-up of 13.8x over SCS and 142.7x over ECOS. \algName~performed no dynamic allocation while SCS and ECOS dynamically allocated the workspace at the beginning due to the use of the CVXPYgen interface, causing them to exceed the total available RAM during execution. Without using the CVXPYgen interface, the dynamically allocated workspace must instead be stored statically, far exceeding the static memory limit. This severely limited SCS and ECOS, with both solvers exceeding total memory limits at $N=64$, while \algName~can scale to $N=256$.


\subsubsection{Early Termination}

High-rate real-time control requires a solver to return a solution within a strict time window. Table~\ref{tab:freq} shows the trajectory-tracking performance of each solver on the rocket soft-landing problem with four different control step durations, resulting in four different time budgets. We solve the same problem as in \ref{ss:rocketSoftLanding}, except that each solver must return within the specified time budget. The maximum number of iterations for each solver was determined based on the average time per iteration for each solver with $N=16$ (Fig.~\ref{fig:rocket_landing}). For example, when given 20ms to solve the problem, the maximum number of solver iterations for ECOS, SCS, and \algName~were 3, 33, and 444, respectively. This represents a factor of 11x to 148x more solver iterations for \algName. Table~\ref{tab:freq} reports two different metrics: A) the total control input violation on box and SOC constraints and B) the landing error (defined as the norm of the deviation between the final and goal states).

ECOS successfully solved to convergence only when given 1000ms, impractical for most real-time control tasks. It failed in subsequent cases due to its limited speed and inability to warm start, with zero iterations completed within 2ms. 
On the other hand, even though SCS and \algName~were both unable to solve the problem to full convergence at every iteration for shorter time budgets, \algName~was able to utilize its increased number of iterations and warm starting to maintain low constraint violation and landing error. This resulted in \algName~outperforming SCS for all scenarios with a 1.6x to 2.4x reduction in landing error and, most critically, while SCS violated constraints across all time budgets, \algName~only appreciably did so for the shortest 2ms time budget.

\begin{table}[t!]
    \centering
    \small
    \vspace{0.5em}
    \caption{Solver performance comparison with different solution-time budgets. Within 20ms ($N$ = 16), the maximum number of solver iterations for ECOS, SCS, and TinyMPC are 3, 33, and 444, respectively. ECOS was not able to complete a single optimization iteration within 2ms.}\label{tab:freq}
    \begin{tabular}{lrrrrr}
        \multicolumn{5}{c}{\textbf{A. Constraint Violation}} \\
        \toprule
        \textbf{Time Budget (ms)} & \textbf{1000} & \textbf{20} & \textbf{10} & \textbf{2} \\ 
        \midrule
        ECOS & \textbf{0.00} & 132.63 & 1757.00 & \textbf{--} \\
        SCS & 2.04 & 5.48 & 10.59 & 22.84 \\
        TinyMPC & 0.01 & \textbf{0.01} & \textbf{0.01} & \textbf{4.43} \\
        \noalign{\vskip 0.5em}
        \multicolumn{5}{c}{\textbf{B. Landing Error}} \\
        \toprule
        ECOS & 1.33 & 629.02 & 939.26 & \textbf{--} \\
        SCS & 1.35 & 1.36 & 1.37 & 2.11 \\
        TinyMPC & \textbf{0.87}  & \textbf{0.87} & \textbf{0.87} & \textbf{0.87} \\
        \bottomrule
    \end{tabular}
    \vspace{-15pt}
\end{table}


\subsection{Robot Hardware Experiments} \label{ss:hardwareExp}
Next, we demonstrate the efficacy of our solver for real-time execution of dynamic control tasks on a Crazyflie 2.1, a 27 gram quadrotor with an ARM Cortex-M4 (STM32F405) clocked at 168 MHz with 192 kB of SRAM and 1 MB of flash. We present three experiments detailing the high performance of \algName~for dynamic control tasks requiring the online solution to QPs and SOCPs: 1)~predictive safety filtering to enable safe control of fundamentally unsafe policies, 2)~attitude/thrust vector regulation with thrust-cone constraints, and 3)~tracking a spiral landing trajectory with conic constraints and a constraint-violating helical reference.


We note that for the problem sizes required for these experiments, OSQP, SCS, and ECOS all could not fit within the memory available on this MCU and, as such, cannot be used as baselines. Instead, we compare against the Brescianini~\cite{bre2013bre} and Mellinger~\cite{Mellinger11} reactive controllers included with the Crazyflie firmware. These controllers often clip the control input to meet hardware constraints. For all experiments, we ran all controllers with their default parameters and attached an optical flow deck to the Crazyflie to perform state estimation fully onboard the robot.

In all experiments, we linearized the quadrotor's 6-DOF dynamics about a hover, representing the quadrotor as a point mass with a thrust vector input, and representing its attitude with a quaternion using the formulation in~\cite{jackson21planning}. This problem has state dimension $n=12$ and $m=4$, representing the quadrotor's full state and PWM motor commands. It is worth noting that the Crazyflie platform offers a great chance to test the controller's robustness due to its high model uncertainty and rapidly depleting battery power (only 5-15 minutes of flight). Under the restricted budget, our \algName\ ran at 50 Hz with at most 20 ADMM iterations per call, using a fast reactive controller to track the predicted next state.

\subsubsection{Predictive Safety Filtering}

We use a nominal PD controller and formulate a predictive safety filtering problem as a QP, similar to~\ref{ss:benchmark}. The Crazyflie was commanded to follow a sinusoidal path along a single axis with an amplitude of 1.2 m (Fig.~\ref{fig:hw_landing_safe} bottom), which was then tracked with both a nominal PD controller (\textcolor{red}{red}) and by \algName~(\textcolor{blue}{blue}) using a horizon of 20 knot points and box constraints at $\pm$0.6 m. The box constraints represent safety limits on the quadrotor's operating space. \algName~is able to successfully respect the safety limits, handling them by slowing to a stop and hovering at the boundaries of the constraints until the reference trajectory comes back around and sends the Crazyflie to the other side of the boundary. This experiment demonstrates \algName's ability to act as a safety layer for unsafe policies.


\begin{figure}[!t]
    \centering
    \vspace{0.5em}
    \begin{subfigure}{\columnwidth}
        \centering
        \resizebox{!}{0.34\columnwidth}{\input{tikz/Attitude_x.tikz}}
    \end{subfigure}\\
    \vspace{0.1em}
    \begin{subfigure}{\columnwidth}
        \centering
        \resizebox{!}{0.405\columnwidth}{\input{tikz/Attitude_y.tikz}}
    \end{subfigure}
    \caption{Attitude/thrust vector regulating performance of different controllers on the Crazyflie. While \algName~was able to constrain the aircraft attitude within the bounds (dashed lines for 0.25 and 0.2 radians), Brescianini and Mellinger exhibited large attitude deviations, causing failures.}
    \label{fig:hw_attitude}
    \vspace{-10pt}
\end{figure}

\subsubsection{Attitude and Thrust-Vector Regulation}

In many controllers for vertical take-off and landing (VTOL) aircraft, the thrust vector is constrained to lie within a cone~\cite{Malyuta2022convex}. We formulated an SOCP-based MPC problem for the Crazyflie that incorporates such a constraint, implicitly constraining the drone's attitude. We used the Brescianini, Mellinger, and \algName~controllers to track an aggressive maneuver (drawing a circle in the air very quickly) to determine if the cone constraint was limiting the Crazyflie's attitude. As depicted in Fig.~\ref{fig:hw_attitude}, \algName~was able to successfully limit the Crazyflie's attitude to two different maximum values (0.2 and 0.25 radians). Conversely, the baselines exhibited significant attitude deviations, resulting in failures. It is important to note that one can only reduce the attitude deviations of these myopic baselines through careful gain tuning, without any guarantees, while \algName~allows them to be specified explicitly as constraints.\footnote{We note that thrust-cone constraints are particularly valuable for \algName~on quadrotors, as the solver relies on linearized dynamics with small-angle approximations, which are only valid within a fixed region of the state space. As such, enforcing a thrust-cone constraint helps ensure that the system remains within this valid operating region, which is essential for maintaining stability during control tasks.}


\subsubsection{Conically Constrained Spiral Landing}

Planetary landing problems typically include a glideslope constraint to ensure sufficient elevation during approach and to prevent the spacecraft from crashing into terrain \cite{Malyuta2022convex}. Fig.~\ref{fig:hw_landing_safe} top demonstrates the ability of \algName~to handle the planetary landing glideslope constraint of a spacecraft. The reference trajectory is a descending cylindrical spiral (\textcolor{red}{red}) which we tracked with \algName~and no position constraints. We then added a conic constraint to restrict the Crazyflie's position to within a 45$^{\circ}$ cone originating from the center of the cylindrical reference trajectory. \algName~restricts the Crazyflie from leaving the cone defined by the glideslope constraint, resulting in a spiral landing maneuver (\textcolor{blue}{blue}).

\section{Conclusions and Future Work} \label{sec:conclusion}
In this paper, we develop \algName, an open-source, high-speed, structure-exploiting, alternating direction method of multipliers (ADMM) solver targeting low-power embedded conic control applications. We also present a code-generation framework with high level \texttt{Python}, \texttt{MATLAB}, and \texttt{Julia} interfaces that makes it easy to use our solver. We demonstrate the performance of \algName~through a series of experiments including a number of microcontroller benchmarks, and hardware deployments using a 27 gram Crazyflie quadrotor~\cite{crazyflie2_1}.

There are several directions for future work. One of particular note is that our approach, like that of~\cite{tinympc}, relies on fixed (set of) linearizations, which may not capture all robotic systems well. To address this, we plan to explore recent work~\cite{sharpless2024state} that models the nonlinear-to-linear gap as an antagonistic disturbance using reachability analysis, enabling us to more safely support nonlinear systems.

\bibliographystyle{styles/IEEEtran}
\bibliography{styles/IEEEabrv,refs.bib}

@article{scs,
    author       = {Brendan O'Donoghue and Eric Chu and Neal Parikh and Stephen Boyd},
    title        = {Conic Optimization via Operator Splitting and Homogeneous Self-Dual Embedding},
    journal      = {Journal of Optimization Theory and Applications},
    month        = {June},
    year         = {2016},
    volume       = {169},
    number       = {3},
    pages        = {1042-1068}
}

@article{frison2020hpipm,
  title={HPIPM: a high-performance quadratic programming framework for model predictive control},
  author={Frison, Gianluca and Diehl, Moritz},
  journal={IFAC-PapersOnLine},
  volume={53},
  number={2},
  pages={6563--6569},
  year={2020},
  publisher={Elsevier}
}

@inproceedings{adabag2024mpcgpu,
  title={MPCGPU: Real-Time Nonlinear Model Predictive Control through Preconditioned Conjugate Gradient on the GPU}, 
  author={Emre Adabag and Miloni Atal and William Gerard and Brian Plancher},
  booktitle={IEEE International Conference on Robotics and Automation (ICRA)},
  address = {Yokohama, Japan},
  month={May.},
  year = {2024}
}

@article{liu2016entry,
  title={Entry trajectory optimization by second-order cone programming},
  author={Liu, Xinfu and Shen, Zuojun and Lu, Ping},
  journal={Journal of Guidance, Control, and Dynamics},
  volume={39},
  number={2},
  pages={227--241},
  year={2016},
  publisher={American Institute of Aeronautics and Astronautics}
}

@article{ben2009robust,
  title={Robust optimization},
  author={Ben-Tal, Aharon and Nemirovski, Arkadi and El Ghaoui, Laurent},
  year={2009},
  publisher={Princeton university press}
}

@article{alizadeh2003second,
  title={Second-order cone programming},
  author={Alizadeh, Farid and Goldfarb, Donald},
  journal={Mathematical programming},
  volume={95},
  number={1},
  pages={3--51},
  year={2003}
}

@article{lobo1998applications,
  title={Applications of second-order cone programming},
  author={Lobo, Miguel Sousa and Vandenberghe, Lieven and Boyd, Stephen and Lebret, Herv{\'e}},
  journal={Linear algebra and its applications},
  volume={284},
  number={1-3},
  pages={193--228},
  year={1998},
  publisher={Elsevier}
}

@inproceedings{Mahajan2025adaptiveTinyMPC,
  title={Robust and Efficient Embedded Convex Optimization through First-Order Adaptive Caching}, 
  author={Ishaan Mahajan and Brian Plancher},
  booktitle={IEEE/RSJ International Conference on Intelligent Robots and Systems (IROS)},
  year={2025}
}

@INPROCEEDINGS{ecos,
author={Domahidi, A. and Chu, E. and Boyd, S.},
booktitle={IEEE European Control Conference (ECC)},
title={{ECOS}: {A}n {SOCP} solver for embedded systems},
year={2013}}

@INPROCEEDINGS{tinympc,
      title={TinyMPC: Model-Predictive Control on Resource-Constrained Microcontrollers}, 
      author={Khai Nguyen and Sam Schoedel and Anoushka Alavilli and Brian Plancher and Zachary Manchester},
      booktitle={IEEE International Conference on Robotics and Automation (ICRA)},
      address = {Yokohama, Japan},
      month={May.},
      year = {2024}
}

@INPROCEEDINGS{reluqp,
      title={ReLU-QP: A GPU-Accelerated Quadratic Programming Solver for Model-Predictive Control}, 
      author={Arun L. Bishop and John Z. Zhang and Swaminathan Gurumurthy and Kevin Tracy and Zachary Manchester},
      booktitle={IEEE International Conference on Robotics and Automation (ICRA)},
      address = {Yokohama, Japan},
      month={May.},
      year = {2024}
}

@online{clarabel,
  author = {Paul Goulart and Yuwen Chen},
  title = {Clarabel},
  year = 2022,
  url = {https://oxfordcontrol.github.io/ClarabelDocs/stable/},
}

@manual{mosek,
   author = "MOSEK ApS",
   title = "Introducing the MOSEK Optimization Suite 10.1.28",
   year = 2024,
   url = "https://docs.mosek.com/latest/intro/index.html"
 }

@misc{forces,
      Author       = "Embotech AG",
      url = "https://forces.embotech.com/",
      Title        = "FORCESPRO",
      Year         = "2014--2023"
}

@inproceedings{altroc,
  title={ALTRO-C: A fast solver for conic model-predictive control},
  author={Jackson, Brian E and Punnoose, Tarun and Neamati, Daniel and Tracy, Kevin and Jitosho, Rianna and Manchester, Zachary},
  booktitle={IEEE International Conference on Robotics and Automation (ICRA)},
  address = {Xi'an, China},
  month={May},
  year={2021}}

@inproceedings{cosmo,
  title={COSMO: A conic operator splitting method for large convex problems},
  author={Garstka, Michael and Cannon, Mark and Goulart, Paul},
  booktitle={IEEE European Control Conference (ECC)},
  year={2019}}

@ARTICLE{megahertz2014,
  author={Jerez, Juan L. and Goulart, Paul J. and Richter, Stefan and Constantinides, George A. and Kerrigan, Eric C. and Morari, Manfred},
  journal={IEEE Transactions on Automatic Control}, 
  title={Embedded Online Optimization for Model Predictive Control at Megahertz Rates}, 
  year={2014},
  volume={59},
  number={12},
  pages={3238-3251},
  doi={10.1109/TAC.2014.2351991}}

@article{Donoghue2013,
author = {O'Donoghue, Brendan and Stathopoulos, Georgios and Boyd, Stephen},
year = {2013},
month = {11},
pages = {2432-2442},
title = {A Splitting Method for Optimal Control},
volume = {21},
journal = {IEEE Transactions on Control Systems Technology}
}

@article{torrente2021data,
  title={Data-driven MPC for quadrotors},
  author={Torrente, Guillem and Kaufmann, Elia and F{\"o}hn, Philipp and Scaramuzza, Davide},
  journal={IEEE Robotics and Automation Letters},
  year={2021}}

@inproceedings{sharpless2024state,
  title={State-Augmented Linear Games with Antagonistic Error for High-Dimensional, Nonlinear Hamilton-Jacobi Reachability},
  author={Sharpless, Will and Chow, Yat Tin and Herbert, Sylvia},
  booktitle={IEEE Conference on Decision and Control (CDC)},
  year={2024}}

@article{lambert2019low,
  title={Low-level control of a quadrotor with deep model-based reinforcement learning},
  author={Lambert, Nathan O and Drew, Daniel S and Yaconelli, Joseph and Levine, Sergey and Calandra, Roberto and Pister, Kristofer SJ},
  journal={IEEE Robotics and Automation Letters},
  year={2019}}

@article{luis2020online,
  title={Online trajectory generation with distributed model predictive control for multi-robot motion planning},
  author={Luis, Carlos E and Vukosavljev, Marijan and Schoellig, Angela P},
  journal={IEEE Robotics and Automation Letters},
  year={2020}}

@article{xi2021gto,
  title={GTO-MPC-based target chasing using a quadrotor in cluttered environments},
  author={Xi, Lele and Wang, Xinyi and Jiao, Lei and Lai, Shupeng and Peng, Zhihong and Chen, Ben M},
  journal={IEEE Transactions on Industrial Electronics},
  volume={69},
  number={6},
  pages={6026--6035},
  year={2021},
  publisher={IEEE}
}

@misc{crazyflie2_1,
  title     = {Crazyflie 2.1},
  author    = {{Bitcraze}},
  url       = {https://www.bitcraze.io/products/crazyflie-2-1/},
  year = {2023}
}

@article{zhang2017visual,
  title={Visual-inertial odometry on chip: An algorithm-and-hardware co-design approach},
  author={Zhang, Zhengdong and Suleiman, Amr AbdulZahir and Carlone, Luca and Sze, Vivienne and Karaman, Sertac},
  year={2017}
}

@article{stellato2020osqp,
  title={OSQP: An operator splitting solver for quadratic programs},
  author={Stellato, Bartolomeo and Banjac, Goran and Goulart, Paul and Bemporad, Alberto and Boyd, Stephen},
  journal={Mathematical Programming Computation},
  volume={12},
  number={4},
  pages={637--672},
  year={2020},
  publisher={Springer}
}

@article{wensing2022optimization,
  author={Wensing, Patrick M. and Posa, Michael and Hu, Yue and Escande, Adrien and Mansard, Nicolas and Prete, Andrea Del},
  journal={IEEE Transactions on Robotics}, 
  title={Optimization-Based Control for Dynamic Legged Robots}, 
  year={2024}}

@article{aydinoglu2023consensus,
  author={Aydinoglu, Alp and Wei, Adam and Huang, Wei-Cheng and Posa, Michael},
  journal={IEEE Transactions on Robotics}, 
  title={Consensus Complementarity Control for Multicontact MPC}, 
  year={2024},
  volume={40},
  number={},
  pages={3879-3896},
  doi={10.1109/TRO.2024.3435423}}

@article{klein1990optimal,
  title={Optimal force distribution for the legs of a walking machine with friction cone constraints},
  author={Klein, Charles A and Kittivatcharapong, Sakon},
  journal={IEEE Transactions on Robotics and Automation},
  year={1990}}

@article{marcucci2020warm,
  title={Warm start of mixed-integer programs for model predictive control of hybrid systems},
  author={Marcucci, Tobia and Tedrake, Russ},
  journal={IEEE Transactions on Automatic Control},
  volume={66},
  number={6},
  pages={2433--2448},
  year={2020},
  publisher={IEEE}
}

@article{le2024fast,
  title={Fast contact-implicit model predictive control},
  author={Le Cleac'h, Simon and Howell, Taylor A and Yang, Shuo and Lee, Chi-Yen and Zhang, John and Bishop, Arun and Schwager, Mac and Manchester, Zachary},
  journal={IEEE Transactions on Robotics},
  year={2024},
  publisher={IEEE}
}

@article{cvxpygen,
  author={Schaller, Maximilian and Banjac, Goran and Diamond, Steven and Agrawal, Akshay and Stellato, Bartolomeo and Boyd, Stephen},
  journal={IEEE Control Systems Letters}, 
  title={Embedded Code Generation With CVXPY}, 
  year={2022},
  volume={6},
  number={},
  pages={2653-2658},
  doi={10.1109/LCSYS.2022.3173209}}

@inproceedings{neuman2022tiny,
  title={Tiny robot learning: challenges and directions for machine learning in resource-constrained robots},
  author={Neuman, Sabrina M and Plancher, Brian and Duisterhof, Bardienus P and Krishnan, Srivatsan and Banbury, Colby and Mazumder, Mark and Prakash, Shvetank and Jabbour, Jason and Faust, Aleksandra and de Croon, Guido CHE and Janapa Reddi, Vijay},
  booktitle={IEEE International Conference on Artificial Intelligence Circuits and Systems (AICAS)},
  year={2022}
}

@article{hsu2023safety,
  title={The safety filter: A unified view of safety-critical control in autonomous systems},
  author={Hsu, Kai-Chieh and Hu, Haimin and Fisac, Jaime F},
  journal={Annual Review of Control, Robotics, and Autonomous Systems},
  volume={7},
  year={2023},
  publisher={Annual Reviews}
}

@inproceedings{bejarano2023multi,
  title={Multi-Step Model Predictive Safety Filters: Reducing Chattering by Increasing the Prediction Horizon},
  author={Bejarano, Federico Pizarro and Brunke, Lukas and Schoellig, Angela P},
  booktitle={IEEE Conference on Decision and Control (CDC)},
  year={2023}}

@inproceedings{Mattingley12,
  title = {{{CVXGEN}}: A Code Generator for Embedded Convex Optimization},
  booktitle = {Optimization {{Engineering}}},
  date = {2012},
  pages = {1--27},
  author = {Mattingley, Jacob and Boyd, Stephen}
}

@article{boyd2011distributed,
  title={Distributed optimization and statistical learning via the alternating direction method of multipliers},
  author={Boyd, Stephen and Parikh, Neal and Chu, Eric and Peleato, Borja and Eckstein, Jonathan and others},
  journal={Foundations and Trends{\textregistered} in Machine learning},
  volume={3},
  number={1},
  pages={1--122},
  year={2011},
  publisher={Now Publishers, Inc.}
}

@book{Boyd04,
  title = {Convex {{Optimization}}},
  publisher = {{Cambridge University Press}},
  date = {2004},
  author = {Boyd, Stephen and Vandenberghe, Lieven}
}

@misc{lewis12optimal,
	author = {Lewis, Frank L. and Vrabie, Draguna and Syrmos, V.L.},
	month = {1},
	title = {{Optimal Control}},
	year = {2012}
}

@book{10,
  location = {{Atlanta, GA}},
  title = {Control {{Model Learning}} for {{Whole}}-{{Body Mobile Manipulation}}},
  date = {2010-07}
}

@inproceedings{Mellinger11,
  title = {Minimum Snap Trajectory Generation and Control for Quadrotors},
  doi = {10.1109/ICRA.2011.5980409},
  booktitle = {{{IEEE International Conference}} on {{Robotics}} and {{Automation}} ({{ICRA}})},
  year={2011},
  author = {Mellinger, D. and Kumar, V.}
}

@ARTICLE{Malyuta2022convex,
  author={Malyuta, Danylo and Reynolds, Taylor P. and Szmuk, Michael and Lew, Thomas and Bonalli, Riccardo and Pavone, Marco and Açıkmeşe, Behçet},
  journal={IEEE Control Systems Magazine}, 
  year={2022},
  volume={42},
  number={5},
  pages={40-113},
  doi={10.1109/MCS.2022.3187542},
  title={Convex Optimization for Trajectory Generation}}

@inproceedings{AMSwarm,
  title={AMSwarm: An Alternating Minimization Approach for Safe Motion Planning of Quadrotor Swarms in Cluttered Environments},
  booktitle = {{{IEEE International Conference}} on {{Robotics}} and {{Automation}} ({{ICRA}})},
  year = {2023},
  author={Adajania, Vivek and Zhou, Siqi and Arun, Singh and Schoellig, Angela}}

@ARTICLE{13rocketlanding,
  author={Açıkmeşe, Behçet and Carson, John M. and Blackmore, Lars},
  journal={IEEE Transactions on Control Systems Technology}, 
  title={Lossless Convexification of Nonconvex Control Bound and Pointing Constraints of the Soft Landing Optimal Control Problem}, 
  year={2013},
  volume={21},
  number={6},
  pages={2104-2113},
  doi={10.1109/TCST.2012.2237346}}

@INPROCEEDINGS{18legged,
  author={Di Carlo, Jared and Wensing, Patrick M. and Katz, Benjamin and Bledt, Gerardo and Kim, Sangbae},
  booktitle={2018 IEEE/RSJ International Conference on Intelligent Robots and Systems (IROS)}, 
  title={Dynamic Locomotion in the MIT Cheetah 3 Through Convex Model-Predictive Control}, 
  year={2018},
  volume={},
  number={},
  pages={1-9},
  doi={10.1109/IROS.2018.8594448}}

@INPROCEEDINGS{18driving,
  author={Babu, Mithun and Oza, Yash and Singh, Arun Kumar and Krishna, K. Madhava and Medasani, Shanti},
  booktitle={IEEE European Control Conference (ECC)}, 
  title={Model Predictive Control for Autonomous Driving Based on Time Scaled Collision Cone}, 
  year={2018}}

@ARTICLE{jackson21planning,
  author={Jackson, Brian E. and Tracy, Kevin and Manchester, Zachary},
  journal={IEEE Robotics and Automation Letters}, 
  title={Planning With Attitude}, 
  year={2021}}

@ARTICLE{bre2013bre,
       title={Nonlinear quadrocopter attitude control},
       author={Brescianini, Dario and Hehn, Markus and D'Andrea, Raffaello},
       year={2013},
       publisher={ETH Zurich}
}

@misc{teensy41,
	title = {{Teensy® 4.1}},
	url = {https://www.pjrc.com/store/teensy41.html},
}

@Article{Verschueren2021,
  Title                    = {acados -- a modular open-source framework for fast embedded optimal control},
  Author                   = {Robin Verschueren and Gianluca Frison and Dimitris Kouzoupis and Jonathan Frey and Niels van Duijkeren and Andrea Zanelli and Branimir Novoselnik and Thivaharan Albin and Rien Quirynen and Moritz Diehl},
  Journal                  = {Mathematical Programming Computation},
  Year                     = {2021},
}

@inproceedings{gurumurthy2025deq,
  title={DEQ-MPC: Deep Equilibrium Model Predictive Control},
  author={Gurumurthy, Swaminathan and Nguyen, Khai and Bishop, Arun L and Kolter, J Zico and Manchester, Zachary},
  booktitle={9th Annual Conference on Robot Learning},
  year={2025}
}

@preamble{ "\ifdefined\DeclarePrefChars\DeclarePrefChars{'’-}\else\fi " }

@book{nesterov1994self,
  title={Self-scaled cones and interior-point methods in nonlinear programming},
  author={Nesterov, Yurii E and Todd, Michael J},
  year={1994},
  publisher={Universit{\'e} Catholique de Louvain. Center for Operations Research and~…}
}
\end{document}